\documentclass[10pt,twocolumn,letterpaper]{article}

\usepackage{cvpr}              % To produce the CAMERA-READY version
\usepackage{graphicx}
\usepackage{amsmath}
\usepackage{amssymb}
\usepackage{booktabs}
\usepackage{multirow}
\usepackage{multicol}
\usepackage{color, colortbl}
\usepackage{pifont}% http://ctan.org/pkg/pifont
\newcommand{\cmark}{\ding{51}}%
\usepackage[pagebackref,breaklinks,colorlinks]{hyperref}

\usepackage[capitalize]{cleveref}
\crefname{section}{Sec.}{Secs.}
\Crefname{section}{Section}{Sections}
\Crefname{table}{Table}{Tables}
\crefname{table}{Tab.}{Tabs.}

\def\cvprPaperID{*****} % *** Enter the CVPR Paper ID here
\def\confName{CVPR}
\def\confYear{2022}

\begin{document}

\title{HOPE: Hierarchical Spatial-temporal Network for Occupancy Flow Prediction }

\author{Yihan Hu, Wenxin Shao, Bo Jiang, Jiajie Chen, Siqi Chai\\
Zhening Yang, Jingyu Qian, Helong Zhou, Qiang Liu\\
Horizon Robotics\\
{\tt\small \{yihan.hu96, nemonswx\}@gmail.com}
}
\maketitle

%%%%%%%%% ABSTRACT
\begin{abstract}
In this report, we introduce our solution to the Occupancy and Flow Prediction challenge in the Waymo Open Dataset Challenges at CVPR 2022. We have developed a novel hierarchical spatial-temporal network featured with spatial-temporal encoders, a multi-scale aggregator enriched with latent variables, and a recursive hierarchical 3D decoder. We use multiple losses including focal loss and modified flow trace loss to efficiently guide the training process. Our method achieves a Flow-Grounded Occupancy AUC of 0.8389 and outperforms all the other teams on the leaderboard.
\end{abstract}

%%%%%%%%% BODY TEXT
\section{Introduction}
\label{sec:intro}
Since 2020, the Waymo Open Dataset~\cite{sun2020scalability,Ettinger_2021_ICCV} has been providing the research communities with high-quality data collected from both LiDAR and camera sensors in real self-driving scenarios, which have enabled a lot of new exciting research. At CVPR 2022, the Waymo Open Dataset Challenge introduces a brand new challenge named ``Occupancy and Flow Prediction'' by expanding original Motion Datasets\cite{Ettinger_2021_ICCV}. This new form of representation~\cite{mahjourian2022occupancy, hu2021fiery} mitigates the shortcomings of the existing representation such as trajectory sets and occupancy. By predicting flow fields and occupancy jointly, the motion and location probability of all agents in the scene can be captured simultaneously. More specifically, given one-second history of agents in a scene, the algorithm is required to forecast all vehicles' occupancy and flow fields in 8 seconds.  The evaluation metric is calculated using Flow-Grounded Occupancy, which measures the quality of predicted occupancy and flow jointly. The algorithm that achieves best AUC score with Flow-Grounded Occupancy wins the challenge. 

In this report, we propose a hierarchical spatial-temporal network named \textit{HOPE}, composed of spatial-temporal encoders, a multi-scale aggregator enriched with latent variables, and a hierarchical 3D recursive decoder. Our model takes the past 10 frames' and the current frame's as input. We enrich the history with a set of rasterized agents' dynamic states with the scene's static information. Multiple modified losses are used to guide the training convergence. For better result, we have also utilized Stochastic Weights Averaging (SWA), test-time augmentation (TTA) and model ensembling techniques to further refine the predictions.

\begin{figure*}[h]
\centering
\includegraphics[width=0.8\textwidth]{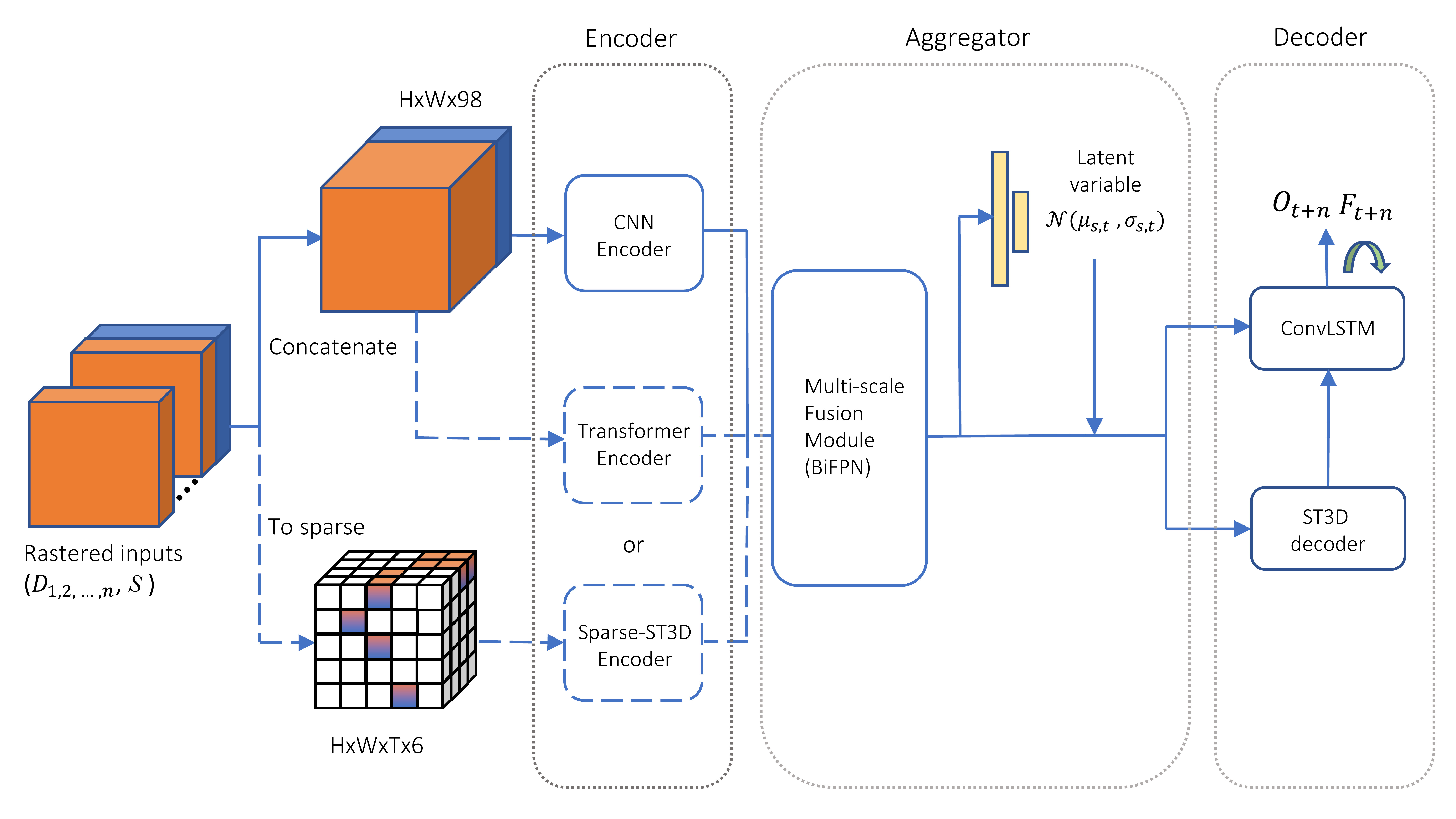} % Reduce the figure size so that it is slightly narrower than the column.
\caption{Overall structure of our model. The inputs include the dynamic features (e.g. agents' occupancy) marked in orange, and the static features (e.g. map states) marked in blue. For the inputs to the 3d sparse encoder, each cube represents both the dynamic features and the static features marked in orange and blue respectively. To reduce the computational cost, we only choose either the transformer encoder or the sparse ST3D encoder to fuse with our CNN encoder.}
\label{fig:framework}
\end{figure*}

\section{Methods}
We divide our model into three parts. The encoder generates multi-scale features from the spatial-temporal inputs. The aggregator interacts and fuses the features at different scales of the encoders. Then, the condensed features are decoded into temporal-coherently predicted waypoints by a spatial-temporal decoder. Fig.~\ref{fig:framework} shows an overview structure of our model. In this section, we first introduce our input representation, and then elaborate the structure of our model.
\subsection{Inputs Representation}
\label{sec:inputs}
For a given scene, our input $I_t = (D_t, S)$ contains both the dynamic features and the static features. The dynamic features $D_t=(O_t, a_t)$ include the occupancy information $O_t$ as well as the attributes of an agent $a_t$ such as bounding box's width, height, length, z , velocity x, velocity y, and speed. Static features $S=s_{1,...,N}$ include road map information and each channel represents a different static element (e.g overall road map, middle line, road edge, traffic light). There are 29 different road features thus $N=28$. We encode past 10 scenes and the current scene for a total of 11 scenes and we use a rasterized bird-eye-view(BEV) as the representation. To be specific, the shape of $O_t$, $a_t$ and $S$ are $H\times W\times 3$(for all 3 classes), $H\times W\times 7$ and $H\times W\times 29$.

For 2D encoders, the dynamic features are concatenated along the temporal dimension, followed by the static features, denoted as $I=(D_{1,...,11}, S)$. We merge the occupancy of bicycle and pedestrian classes, thus the channel of occupancy is 22. For agent attributes, since bounding box's width, height and length features are also static, we merge them across time, thus the channel of agent attributes is $4\times11+3=47$. With 29 channels of road features $S$, the total input channel is 98.

For 3D encoder, we encode the time as an additional dimension, as $I=I_{1,...,11}$ for both the dynamic features and the static features, which is achieved by duplicating the static features across the time dimension. We also simplify both the dynamic features and the static features: for the dynamic features, only the occupancy information of the 3 classes is encoded; and for the static features, we merge all road map information into 3 channels: overall road center, overall road edge, and all other road information (e.g. stop sign, crosswalk, and speed bump). Thus, each feature is 6 dimensional and the final input shape before sparsify is $H\times W\times T\times 6$.

% Our Models take rasterized bird-eye-view(BEV) features as input. We generate two kinds of input for 2D encoder and 3D encoder respectfully. For 2D encoder, temporal information is encoded into channels. For the dynamic objects, we encode the objects' features as occupancy and dynamic attributes (i.e bbox width, height, length, z , velocity x, velocity y and speed) for the both past and current states. For the static information, we render the map on BEV as one-hot vectors, and each channel represents a different static element(e.g road line, road edge). The dynamic and static features are then concatenated as final feature map. For 3D encoder, we encode the time as an additional dimension. For each feature at a BEV location at time t, there are a total of 6 values. The first 3 are occupancy information of 3 classes and the last 3 are road center, road edge, and other road information respectfully.
\subsection{Model Structure}
\subsubsection{Encoder}
We design and combine 3 different encoders for the task: a 2D CNN encoder, a Transformer-based encoder, and a spatial-temporal sparse 3D CNN encoder. Then the encoded features by different encoders are concatenated and fused at each scale.

\textbf{CNN Encoder}
We use RegNet~\cite{radosavovic2020designing} as our 2D CNN encoder. RegNet is a well-designed backbone featured with high efficiency. In our implementation, we have utilized RegNetX-32GF. The encoder generates multi-scale features $C_1, C_2, C_3, C_4, C_5$, where $C_i$ denotes the feature of spatial size $\frac{H}{2^i} \times \frac{W}{2^i}$ as the input of the aggregator. 

\textbf{SwinTransformer}
\label{sec:swin}
Recently, Transformer-based models have achieved great success on vision tasks. Specifically, the SwinTransformer~\cite{liu2021swin, liu2021swinv2} has become one of the state-of-the-art models on many dense prediction tasks such as detection and segmentation.  The SwinTransformer is an excellent choice for our task for three reasons: First, the occupancy flow prediction is also a dense prediction task; Second, the attention mechanism is widely used in motion prediction tasks ~\cite{deo2021multimodal,ngiam2021scene,gao2020vectornet,zhao2020tnt} as it matches well with the real-world driving scenarios, where interactions occur between agents as well as between agents and roads. Especially, the window attention mechanism greatly reduces the computational cost when the attention ranges increase. Third, the huge amount of data by Waymo Open Dataset provides an excellent and necessary condition for the SwinTransformer to converge.

\begin{figure}[h]
\centering
\includegraphics[width=0.45\textwidth]{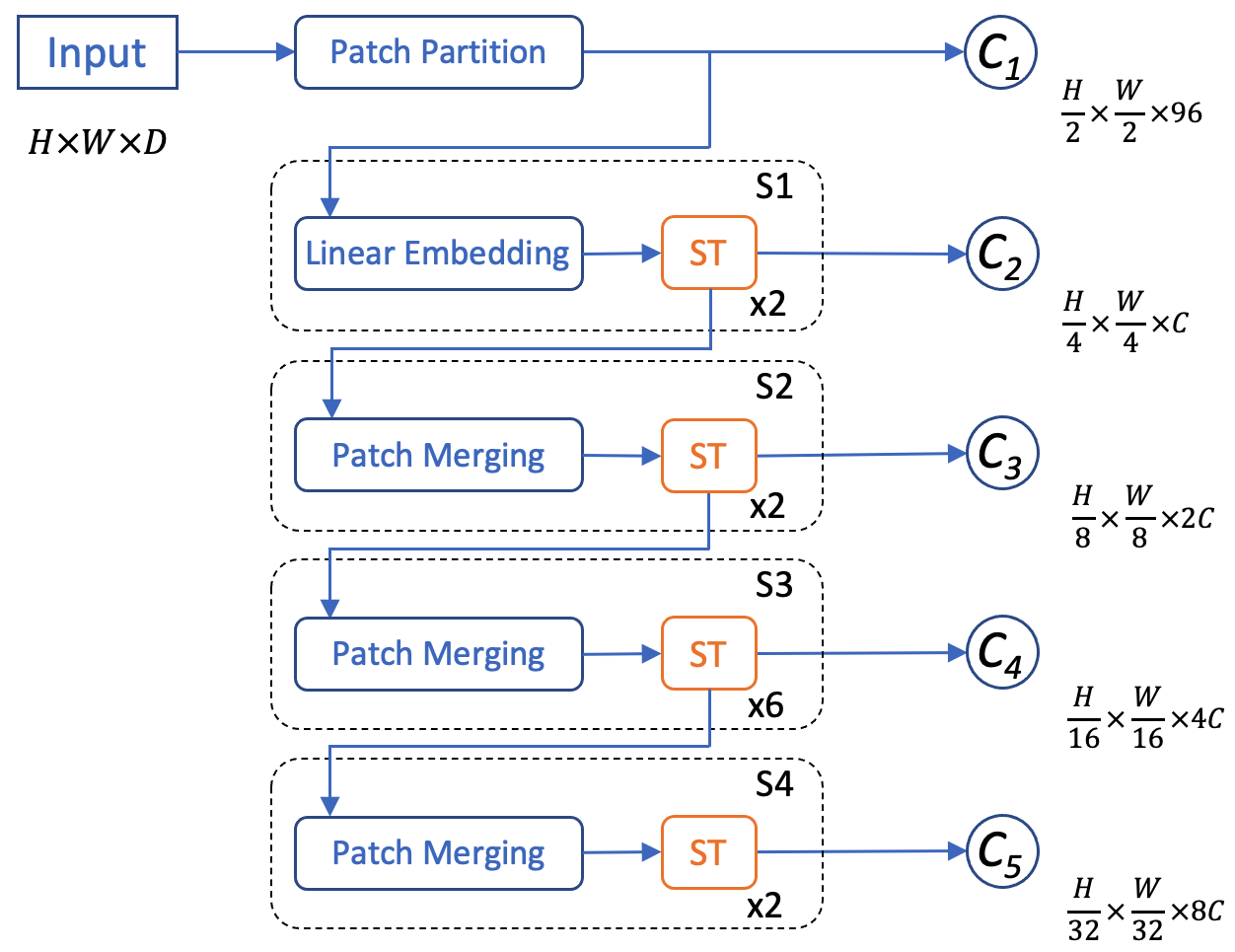}
\caption{The structure of the SwinTransformer encoder. `ST' stands for Swin Transformer Block. $H, W, D, C$ are the input height, width, channel and embedding channel, respectively. $C_1$-$C_5$ are the generated multi-scale features.}
\label{fig:swin}
\end{figure}

In practice, we use SwinV2-T as one of the encoder networks. Similar as the CNN encoder, it also generates multi-scale features $C_1, C_2, C_3, C_4, C_5$, thus these features can be concatenated with the features from other encoders of the same spatial sizes. Note that we use the ImageNet1K~\cite{deng2009imagenet} pretrained model with a window size of 16, and we only change the patch size from 4 to 2 and leave other hyper parameters unchanged compared with the original SwinV2-T. Figure \ref{fig:swin} is an overview of the SwinTransformer module in our task.

\begin{figure}[h]
\centering
\includegraphics[width=0.45\textwidth]{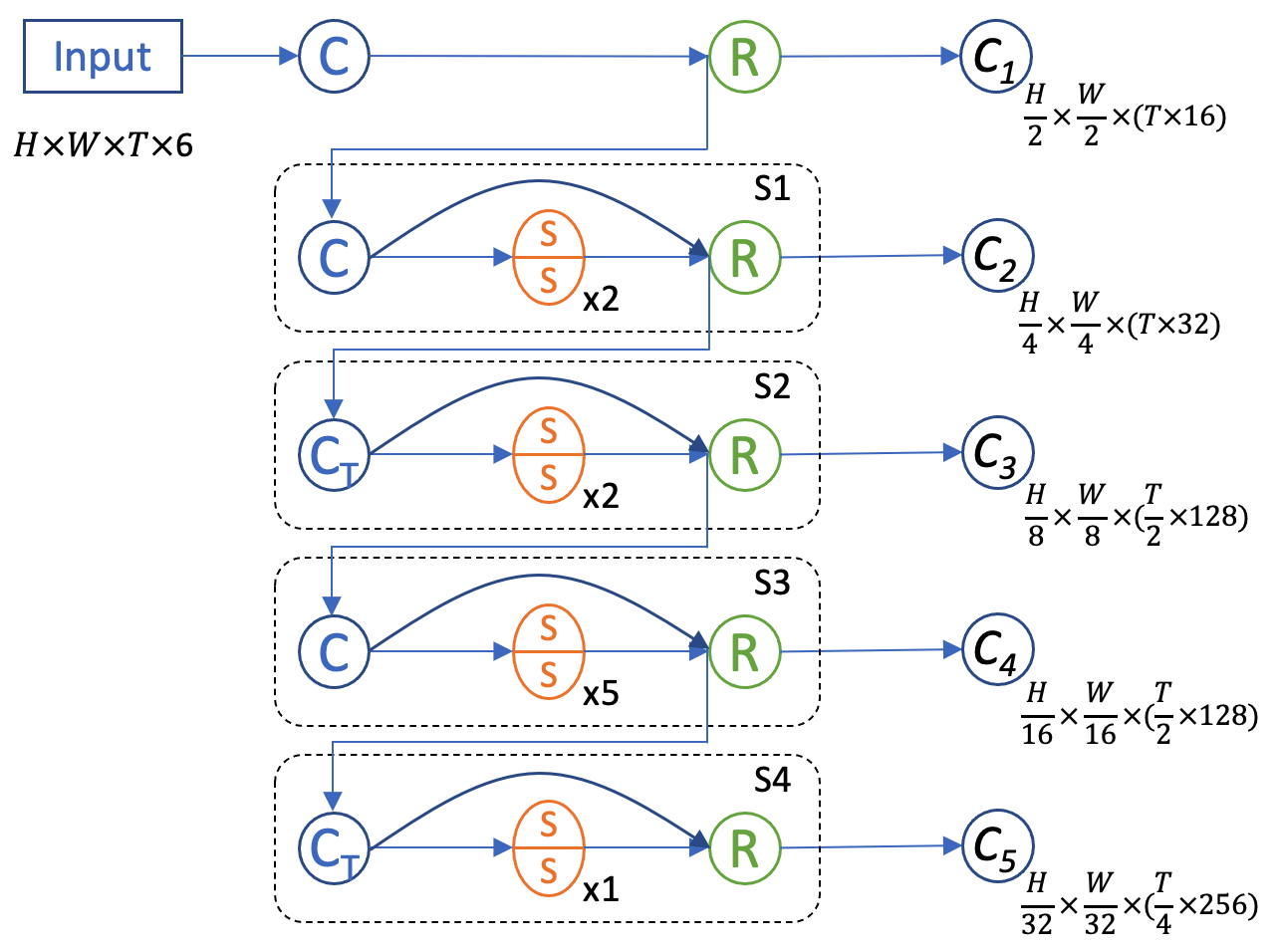} % Reduce the figure size so that it is slightly narrower than the column.
\caption{The structure of the 5-stage 3D encoder. `C' represents 3D sparse convolutional layer that downsamples H and W dimension, `C\textsubscript{T}' represents 3D sparse convolutional layer that downsamples H, W, and T dimension, `S' represents submanifold sparse convolutional layer (two stacked `S' means two such layers in a row), and `R' represents reshape operation that merges time dimension and feature dimension.}
\label{fig:sparse}
\end{figure}

\textbf{Spatial-temporal 3D Encoder}
\label{sec:ST3D}
To better exploit the temporal or inter-frame information, we have designed a spatial-temporal 3D encoder to capture temporal consistency. Since the rasterized input data has high sparsity, we have used 3D sparse convolutional layer and submanifold sparse convolutional layer~\cite{graham2017submanifold} to both accelerate the forward speed and better capture the input features across the spatial and temporal dimensions. We construct a 5-stage backbone, where each stage downsamples the spatial dimensions by 2. For the temporal dimension, it is only downsampled at stage 2 and 4. The output of each stage is gathered and the temporal and feature dimensions of each output are merged as the final feature dimension. Figure \ref{fig:sparse} shows the detailed structure of the sparse encoder.

\subsubsection{Aggregator}
\begin{figure}[h]
\centering
\includegraphics[width=0.45\textwidth]{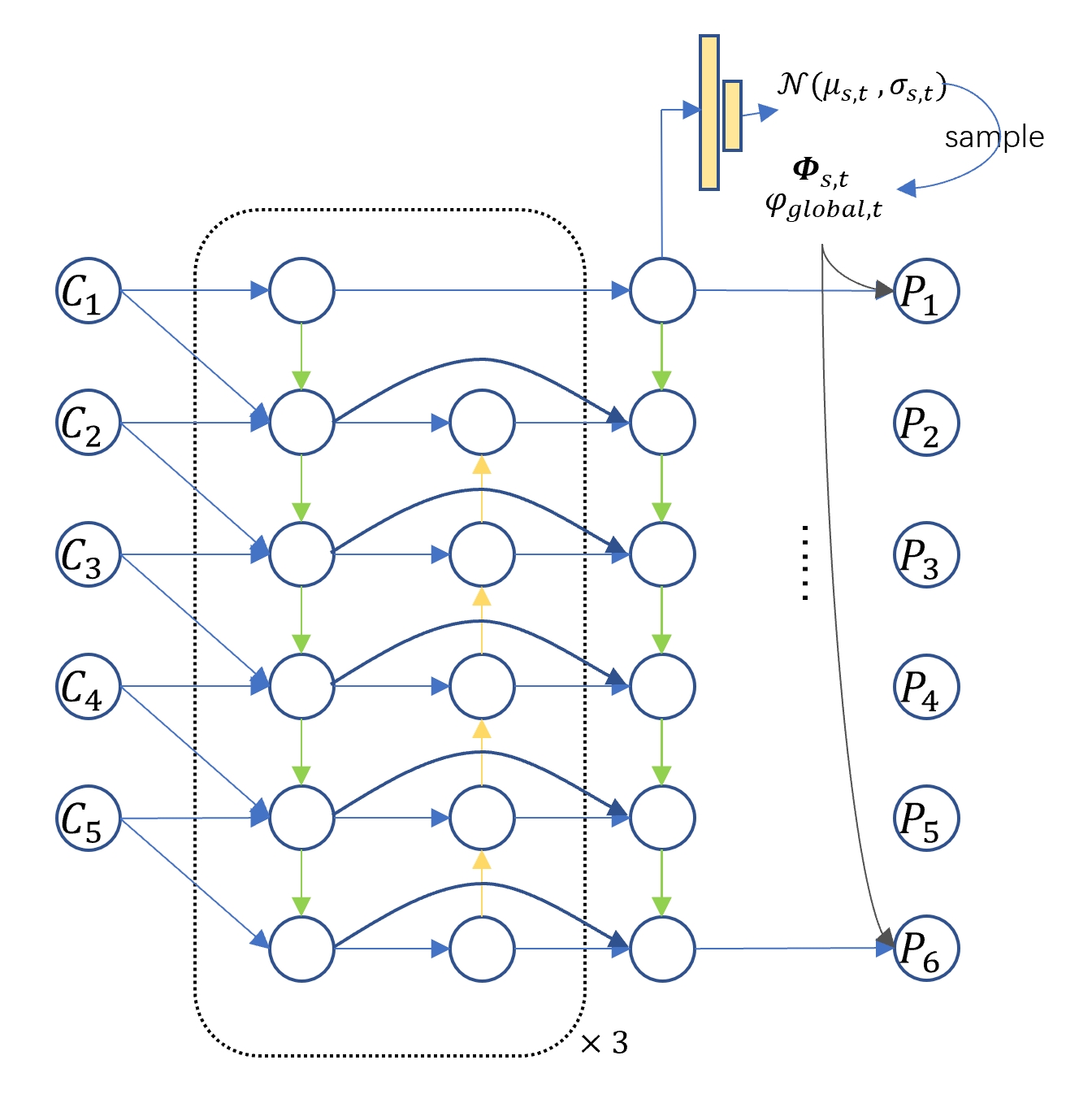} % Reduce the figure size so that it is slightly narrower than the column.
\caption{The structure of the aggregator. $C_1$, ..., $C_5$ are the fused features from 5 different scales. $\Phi_{s,t}$ and $\phi_{global, t}$ are the latent BEV map and the global latent vector respectively, which are sampled from the predicted distributions at scale s. The latent map are then concatenated with the BiFPN features as the aggregated output $P_1$, ..., $P_6$.}
\label{fig:aggregator}
\end{figure}

As shown in Fig.~\ref{fig:aggregator}, the aggregator is composed of a set of stacked BiFPN layers and a multi-scale latent variable module, which aggregates the spatial features at different scales and fuses them with the latent information.

\textbf{BiFPN}
Following EfficientDet~\cite{tan2020efficientdet, mahjourian2022occupancy}, the multi-scale, temporal-condensed features are fused and interacted in a bidirectional manner using BiFPN layers. 

\textbf{Latent Variable} Inspired by FIERY~\cite{hu2021fiery}, we model the inherent stochastic future using latent variables by a conditional variational approach. However, the latent code in FIERY only models the entire scene, mainly the ego vehicle, which is not enough for our task. To expand it to all of the agents in the scene, we model the latent variables as Gaussian distributions in multi-scale bird-eye-view(BEV). During training, we sample the latent code from the current distribution at every location and scale, and take the mean of the distribution during inference. The latent feature maps ($\Phi_{s,t}$ with the shape $H_s\times W_s\times C$) and the vector ($\phi_{global,t}$ with the shape $1\times 1\times C$) are then expanded, concatenated and fused with the multi-scale features. Also, to enforce the consistency of the predicted and observed future distribution, the Kullback-Leibler divergence loss is adopted during training as described in ~\ref{loss:kl}

\subsubsection{Decoder}
\label{sec:decoder}
\begin{figure}[h]
\centering
\includegraphics[width=0.45\textwidth]{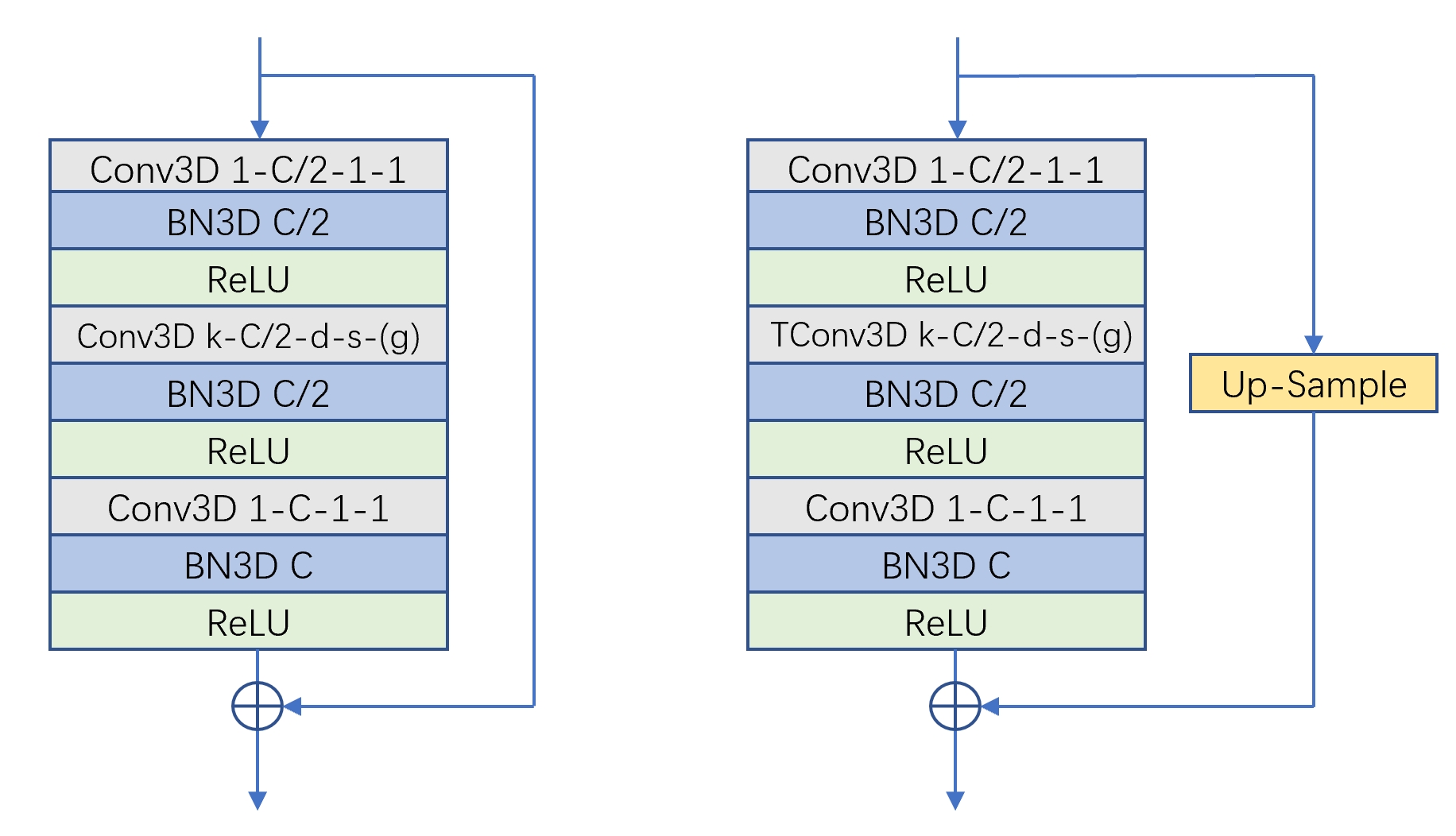} % Reduce the figure size so that it is slightly narrower than the column.
\caption{The structure of the 3D convolutional bottleneck block(Conv3D-BN). ``Conv3D" stands for 3D convolutional layer and ``TConv3D" stands for transpose 3D convolutional layer. The format of the layer setting follows ``kernel size-channels-dilations-strides-(groups)", \textit{i.e.} $k$-$C$-$d$-$s$-($g$).  The kernel size, dilations, and strides are tuples following (T, H, W), where T represents the time domain, and H, W represent the spatial domain. To match the output shape of transpose conv3d layer, a temporal up-sample layer is used at the skip connection.}
\label{fig:bottleneck}
\end{figure}

\begin{figure}[h]
\centering
\includegraphics[width=0.45\textwidth]{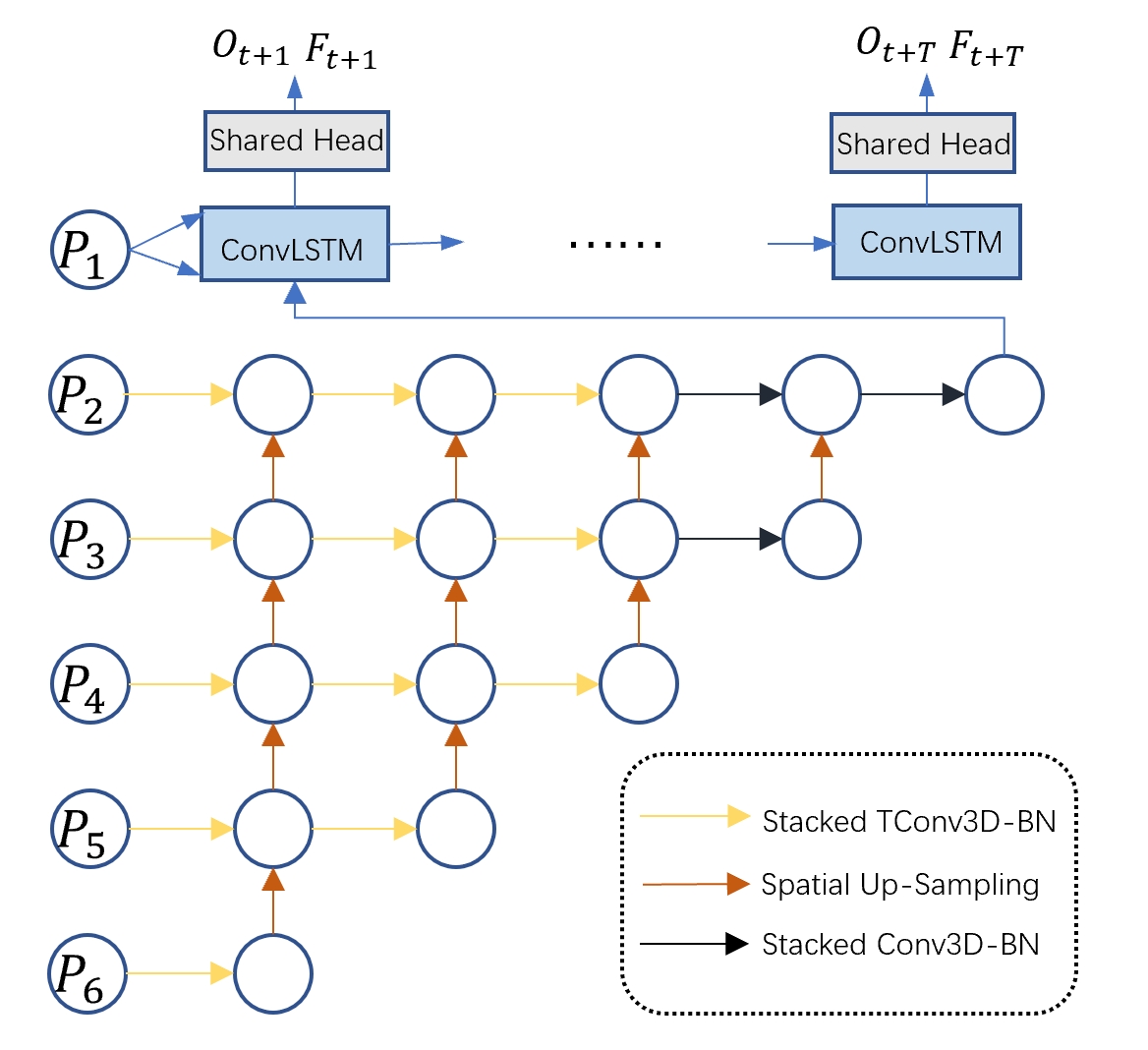} % Reduce the figure size so that it is slightly narrower than the column.
\caption{The structure of the Spatial Temporal Decoder. ``Conv3D-BN" and ``TConv3D-BN" stand for the regular and transposed 3D convolutional bottleneck block respectively. }
\label{fig:decoder}
\end{figure}

Different from image segmentation task which only considers spatial information, the occupancy and flow prediction take both of the spatial and temporal information into account. Thus, naively applying the decoders from segmentation task is not the optimal choice. Recently, spatial-temporal prediction has gained more advancements, such as in video prediction~\cite{oprea2020review}. Inspired by the previous work~\cite{shi2015convolutional, wang2018eidetic,hu2019learning,hu2020probabilistic, hu2021fiery}, we design a novel hierarchical spatial-temporal decoder as shown in Fig.~\ref{fig:decoder}. Different from the previous video prediction task, the occupancy and flow prediction has much larger spatial dimension (256x256) and a longer forecasting time horizon (8 second), which implies long range dependence in both spatial and temporal domain. Thus, widely used ConvGRU~\cite{ballas2015delving} and ConvLSTM~\cite{shi2015convolutional} are not an optimal choice since they have small receptive field and relatively large computational cost due to their recursive nature. To enlarge our receptive field without sacrificing the efficiency, we design stacked dilated 3D convolutional bottleneck structures as shown in Fig.~\ref{fig:bottleneck}. The dilation~\cite{chen2017deeplab} makes our receptive field extremely large with almost the same computational complexity. Also, the grouped bottleneck structures with skip connections further boost our model's speed and efficiency. To incorporate multi-scale features, the stacked bottlenecks are then combined in a FPN manner as shown in Fig.~\ref{fig:decoder}. Since our input multi-scale features from the aggregator are condensed in temporal domain (i.e $P_2$,  $P_3$, ..., $P_6$), we further adopt transposed 3D convolutions in the bottleneck structures which gradually unroll the temporal domain. 

To incorporate fine-grained features efficiently, we adopt a single ConvLSTM layer at the top. As shown in Fig.~\ref{fig:decoder}, the ConvLSTM layer takes largest scale output by the aggregator (i.e $P_1$) as the hidden and the cell states. As a refinement, the ConvLSTM layer recursively takes the temporal-unrolled coarse output by the 3D-FPN structures as input. Then, the unrolled features are sent to the shared heads which are finally decoded as the observed/occluded occupancy (${O}_{t+n}$) and the flow prediction (${F}_{t+n}$). 

\subsection{Losses}
\textbf{Occupancy Loss} Focal loss~\cite{lin2017focal} is adopted to supervise both the observed and occluded occupancy prediction, as shown in Eq.~\ref{eq:occ}. 
\begin{equation}
\label{eq:occ}
    \mathcal{L}_{O}^{b,c}=\sum_{t=1}^{T_{\text {pred }}} \sum_{x=0}^{w-1} \sum_{y=0}^{h-1} \mathcal{F}\left(O_{t}^{b,c}(x, y), \tilde{O}_{t}^{b,c}(x, y\right)
\end{equation}
\begin{equation}
    \mathcal{L}_{occ}=  \lambda_{O}^b\mathcal{L}_{O}^{b} + \lambda_{O}^c\mathcal{L}_{O}^{c}
\end{equation}
where $O^b$ is observed occupancy, $O^c$ is occluded occupancy, $O_t$ is the predicted occupancy and $\tilde{O}_{t}$ is the ground truth occupancy at time step $t$. 

\textbf{Flow Loss} Similar to Mahjourian et al.~\cite{mahjourian2022occupancy}, we apply smooth L1 loss for the flow regression. To decouple the flow prediction and occupancy prediction tasks, the loss is further weighted by the ground truth occupancy $\tilde{O}_{t}$.

\textbf{Traced Loss} We implement the traced loss as described in Mahjourian et al.~\cite{mahjourian2022occupancy} with some modifications. Instead of using cross-entropy loss alone, we mix the focal loss and the cross-entrophy loss to further boost the Flow-Grounded Occupancy AUC metric. Also, to be consistent with the Flow-Grounded Occupancy AUC metric, instead of recursively applying the warping process on current occupancy $O_0$, we directly warp the ground truth $O_{t-1}$ at the previous time step with the predicted flow $F_t$ , as shown in Eq.~\ref{eq:warp}.

\begin{equation}
\label{eq:warp}
\mathcal{W}_{t}=F_{t} \circ \tilde{O}_{t-1}
\end{equation}

Then cross entropy loss and focal loss are applied jointly as the final traced loss, as shown in Eq.~\ref{eq:traced}
\begin{equation}
\label{eq:traced}
\begin{aligned}
 \mathcal{L}_\text{traced}=\sum_{t=1}^{T_{\text {pred }}} \sum_{x=0}^{w-1} \sum_{y=0}^{h-1} \lambda_{ce}\mathcal{H}\left(\mathcal{W}_{t}(x, y) O_{t}(x, y), \tilde{O}_{t}(x, y)\right) + \\ \lambda_{f}\mathcal{F}\left(\mathcal{W}_{t}(x, y) O_{t}(x, y), \tilde{O}_{t}(x, y)\right)   
\end{aligned}
\end{equation}
where $\mathcal{H}$ is the cross entropy loss.

\textbf{Probabilistic Loss}
\label{loss:kl}
Following FIERY~\cite{hu2021fiery}, we use Kullback-Leibler divergence to enforce the consistency between the predicted future distributions and the observed future distributions. Different from FIERY, we average the KL-divergence loss over every pixel and scale, as shown in Eq.~\ref{eq:kl} and Eq.~\ref{eq:kll}.
\begin{equation}
\label{eq:kl}
    L_{\text {prob}}=\frac{1}{Nhw}\sum_{n=0}^{N}\sum_{x=0}^{w-1} \sum_{y=0}^{h-1} \lambda_{n} D_n
\end{equation}

\begin{equation}
\label{eq:kll}
    D_n=D_{\mathrm{KL}}\left(F\left(\cdot \mid s_{n,t}, y_{t+1}, \ldots, y_{t+T}\right) \| P\left(\cdot \mid s_{n,t}\right)\right)
\end{equation}
where N is number of scales, $s_{n,t}$ is the n-th output feature map, $y_{t+1}, ..., y_{t+T}$ are ground truth occupancy and flows. In our experiment, we set equal weight to each scale, i.e $\lambda_{s}=1$.

We use the weighted sum of all losses as the final loss:
\begin{equation}
\mathcal{L} = \lambda_\text{occ}\mathcal{L_\text{occ}} + \lambda_\text{flow}\mathcal{L_\text{flow}} + \lambda_\text{traced}\mathcal{L_\text{traced}} + \lambda_\text{prob}\mathcal{L_\text{prob}}
\end{equation}

\section{Experiments}
\subsection{Experiment Settings}
We generate the rasterize input within a $120m\times 120m$ region at a resolution of 0.156m/pixel, thus the input spatial size is $768\times 768$. To match the output resolution with the evaluation range, the mutliscale feature maps are cropped with a 2/3 ratio to keep the $80m\times 80m$ region before the decoder. We use AdamW optimizer~\cite{loshchilov2017decoupled} and cosine annealing policy~\cite{loshchilov2016sgdr} with an initial learning rate of $2.5\times 10^{-4}$. The weight decay is set to 0.01. We set the loss coefficient $\lambda_\text{occ}=500$, $\lambda_\text{flow}=1$, $\lambda_\text{traced}=500$ and $\lambda_\text{prob}=1$. For the occupancy loss, weights between the observed and occluded occupancy are equal, i.e  $\lambda_{O}^{b}=\lambda_{O}^{c}=1$. For the traced loss, weights between the cross entropy and the focal loss are equal, i.e  $\lambda_{ce}=\lambda_{f}=1$. 

\textbf{Model Variations and Ensembling}
We train two model variants: HOPE-Swin and HOPE-3D. HOPE-Swin fuses RegNet and SwinV2 encoder while HOPE-3D fuses RegNet and Sparse encoder. Both models are trained on the training and validation datasets. HOPE-Swin and HOPE-3D are trained for 4 and 2 epochs respectively.

For both models, we use Stochastic Weights Averaging (SWA) ~\cite{izmailov2018averaging,hu2021afdetv2} to further enhance the training. We train each model for one additional epoch using decreased learning rate and greedily average the weights.

\subsection{Test-Time Augmentation and Model Ensembling}
We also boost predictions through Test-Time Augmentation (TTA)~\cite{Krizhevsky2017tta, ding20201st}, by rotating the raster inputs by \(\pi\), and ensembling by taking the weighted average with the original predictions. Note that we only apply TTA on flow predictions, keeping occupancy predictions as the original. The weight of the TTA prediction is 0.25. The effectiveness of TTA is shown in Tab.~\ref{tbl:ablation study}.

After the TTA, we ensemble the output of the two model variants by taking a weighted averaging as the final result.

\section{Results}
\subsection{Performance on the Test Set}
\begin{figure*}
\begin{center}

\includegraphics[width=1\textwidth]{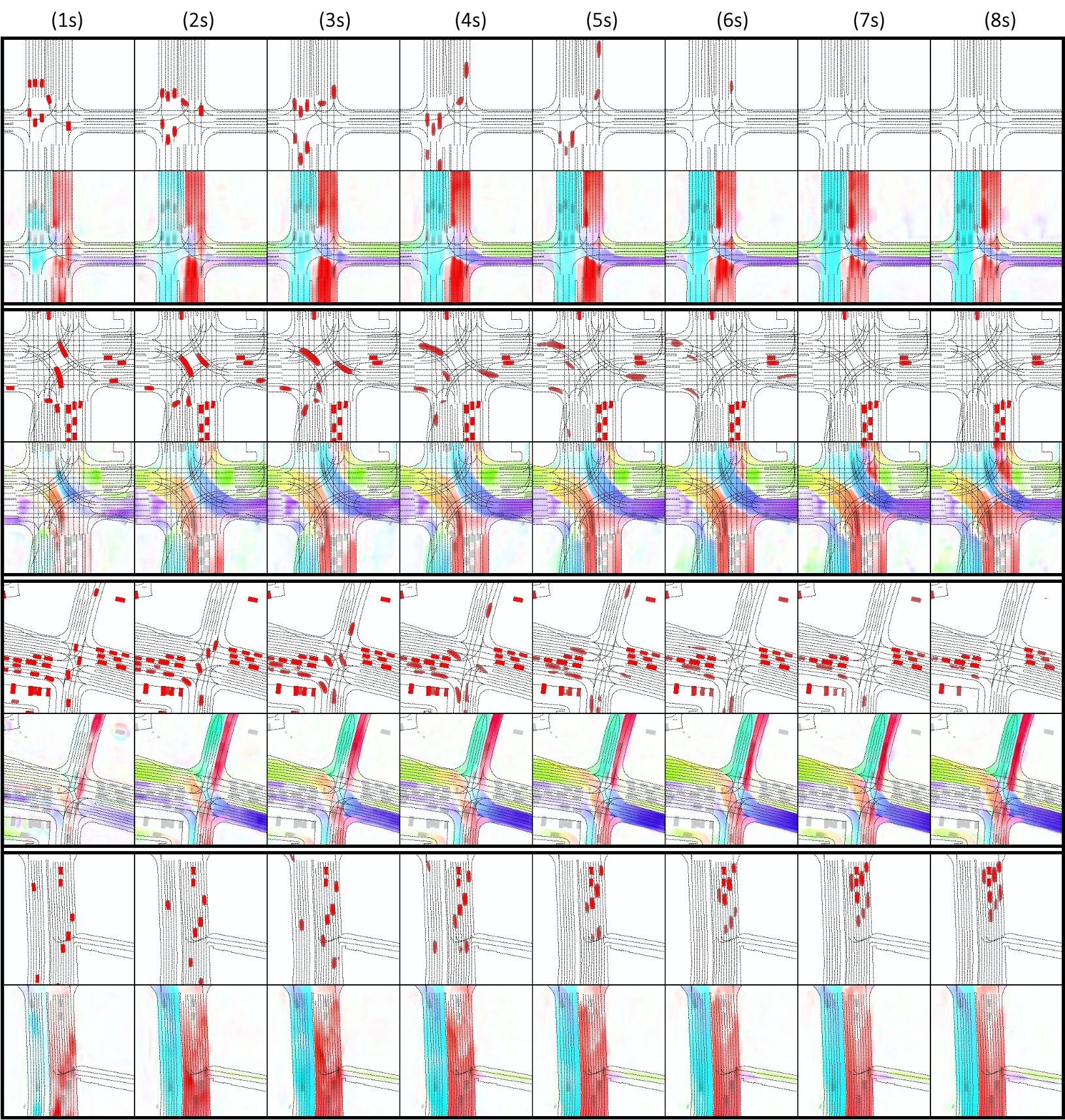}
\end{center}
\caption{Prediction visualization of 4 scenes. For each scene, the first row shows the predicted occupancy with score threshold of 0.4 and the second row shows the predicted flow.}
\label{fig:result vis}
\end{figure*}

The final results on the official Occupancy and Flow Prediction challenge leaderboard are shown in Tab.~\ref{tbl:OccFlow result}. Based on the primary metric, Flow-Grounded Occupancy AUC, we outperform all the other teams. Fig.~\ref{fig:result vis} shows the visualization of some of our predictions.
\begin{table*}[h]
\begin{center}
\definecolor{Gray}{gray}{0.9}
\newcolumntype{g}{>{\columncolor{Gray}}c}
\resizebox{1\textwidth}{!}{%
\begin{tabular}{l|c c|c c|c|g c}
\hline
 \multirow{2}{*}{Models}  &\multicolumn{2}{c|}{Observed Occupancy} &\multicolumn{2}{c|}{Occluded Occupancy} &\multicolumn{1}{c|}{Flow} &\multicolumn{2}{c}{Flow-Grounded Occupancy}\\
  &AUC &Soft IoU &AUC &Soft IoU &EPE &AUC &Soft IoU\\
\hline
\hline
HorizonOccFlowPrediction(Ours) &\textbf{0.8033} &0.2349 &0.1650 &0.0169 &3.6717 &\textbf{0.8389} &\textbf{0.6328}\\
Look Around &0.8014 &0.2336 &0.1386 &0.0285 &\textbf{2.6191} &0.8246 &0.5488\\
Temporal Query - Stable &0.7565 &0.3934 &0.1707 &0.0404 &3.3075 &0.7784 &0.4654\\
STrajNet &0.7514 &0.4818 &0.1610 &0.0183 &3.5867 &0.7772 &0.5551\\
VectorFlow &0.7548 &\textbf{0.4884} &\textbf{0.1736} &\textbf{0.0448} &3.5827 &0.7669 &0.5298\\
\hline
\end{tabular}
}
\end{center}
\caption{Top five submissions of the Occupancy and Flow Prediction challenge. For each team, only the best submission is kept. The results are evaluated on the test set. The Flow-Grounded Occupancy AUC (marked in grey) is used as primary metric.} 
\label{tbl:OccFlow result}
\end{table*}

% Table for Ablation Study
\begin{table*}[h]
\begin{center}
    \definecolor{Gray}{gray}{0.9}
    \newcolumntype{g}{>{\columncolor{Gray}}c}
    \resizebox{\textwidth}{!}{
        \begin{tabular}{cccccccc|gc|cc|c}
        \hline
          %\multirow{2}{*}{$\mathcal{F}$}%
          focal & traced & enriched  & latent  & ST  & 
         +swin  & +ST3D  & \multirow{2}{*}{TTA}& \multicolumn{2}{c|} {Flow-Grounded}  & \multicolumn{2}{c|} {Observed Occupancy}  & Flow \\
         loss & loss & inputs & var. & decoder & encoder & encoder& & AUC & Soft-IoU & AUC & Soft-IoU & EPE\\
        \hline
         & & & & & & & & 0.7492 & 0.4905 & 0.7193 & 0.4403 & 4.4692
        \\
         \cmark & & & & & & & & 0.7693 & 0.5109 & 0.7133 & 0.1924 & 4.2966\\
         \cmark& \cmark& & & & & & & 0.7739 & 0.5344 & 0.6792 & 0.1503 & 4.7153\\
         \cmark& \cmark&\cmark & & & & & &0.7862 & 0.5511 & 0.7053 & 0.1610 & 4.3068\\
         \cmark& \cmark& \cmark& \cmark& & & & & 0.7850 & 0.5523 & 0.7065 & 0.1655 & 4.3146\\
         \cmark& \cmark& \cmark& \cmark& \cmark & & & & 0.7956 & 0.5580 & 0.7231 & 0.1466 & 4.1320\\
         \cmark& \cmark& \cmark& \cmark&\cmark &\cmark & & & 0.7972 & 0.5627 & 0.7258 & 0.1619 & 4.0951\\
         \cmark& \cmark& \cmark& \cmark&\cmark & & \cmark& & 0.7983 & 0.5647 & 0.7262 & 0.1611 & 4.0433\\
         \cmark& \cmark& \cmark& \cmark&\cmark & & \cmark& \cmark& 0.8023 & 0.5788 & 0.7262 & 0.1611 & 3.9840\\
        \hline
        \end{tabular}
        }
\end{center}
\caption{The detailed ablation study. All the experiments are conducted on a 1/10 training set for 5 epochs and evaluated on the all validation set with the scenario IDs provided by WOD. The rasterized inputs are fixed at a range of 80mx80m. The metrics are evaluated under our customized environment, which is slightly different from the online evaluation server. For WOD2022 challenge, the ``Flow-Grounded AUC'' (marked in grey) is used as primary metric.} 
\label{tbl:ablation study}
\end{table*}

\subsection{Ablation Studies}
To verify the effectiveness of our approach, we have conducted a number of ablation studies as shown in Fig.~\ref{tbl:ablation study}. 

\textbf{Losses} We replace the losses in our baseline model with the focal loss and the traced loss. The Flow-Grounded Occupancy AUC and Flow-Grounded Occupancy soft-IoU are boosted dramatically by replacing the cross-entropy loss with the focal loss~\cite{lin2017focal}, at some decrease of the Observed soft-IoU metric. Moreover, adding traced loss~\cite{mahjourian2022occupancy} further boosts Flow-Grounded Occupancy AUC and Flow-Grounded Occupancy soft-IoU by a large margin.

\textbf{Additional Input} As described in ~\ref{sec:inputs}, besides the simple rasterized occupancy $O_t$, we further enrich the dynamic features $D_t=(O_t, a_t)$ by rasterizing the agents' attribute $a_t$(i.e bounding box's width, height, length, z, velocity x, velocity y and speed). Also, we expand the static feature $S_t$ from a single one-hot layer to 28 layers, which includes lanes' and traffic states' information. By adding additional information, we observed significant improvements among all metrics. 

\textbf{Probabilistic Modeling} Adding multi-scale latent variable slightly improves the Flow-Grounded Occupancy soft-IoU, Observed AUC, and Observed soft-IoU with the cost of a slight drop in Flow-Grounded Occupancy AUC.

\textbf{Spatial Temporal Decoder} Instead of predicting all 8 waypoints at the same time, we add a spatial temporal decoder and predict 8 future states recursively as described in ~\ref{sec:decoder}. We have observed significant improvements in all metrics, showing the effectiveness of decoupling the spatial and temporal information when decoding.

\textbf{Extra Encoder} To further enhance our encoded features, we have designed two variant models: One by adding SwinTransformer encoder~\ref{sec:swin} and the other by adding ST3D encoder ~\ref{sec:ST3D}. We can observe in Tab.~\ref{tbl:ablation study} that fusing both encoders outperforms the single CNN encoder. Moreover, we can further benefit from the two variants of the model by using model ensemble technique.

\section{Conclusion}
In this report, we have shown our method for the occupancy flow prediction with a hierarchical spatial-temporal model named \textit{HOPE}. This model is composed of multiple spatial-temporal encoders, a multi-scale latent code enriched aggregator, and a hierarchical 3D recursive decoder. In addition, we have redesigned multiple losses, including the focal loss and the flow trace loss. For further enhancements, we have adopted various ensemble techniques, i.e stochastic weights averaging and test time augmentation. Thanks to all these innovations, \textit{HOPE} achieved the top accuracy in term of the Flow-Grounded Occupancy AUC for the ``Occupancy and Flow Prediction" in the Waymo Open Dataset Challenges 2022.

%%%%%%%%% REFERENCES
{\small
\bibliographystyle{ieee_fullname}
\bibliography{egbib}

\begin{thebibliography}{10}\itemsep=-1pt

\bibitem{ballas2015delving}
Nicolas Ballas, Li Yao, Chris Pal, and Aaron Courville.
\newblock Delving deeper into convolutional networks for learning video
  representations.
\newblock {\em arXiv preprint arXiv:1511.06432}, 2015.

\bibitem{chen2017deeplab}
Liang-Chieh Chen, George Papandreou, Iasonas Kokkinos, Kevin Murphy, and Alan~L
  Yuille.
\newblock Deeplab: Semantic image segmentation with deep convolutional nets,
  atrous convolution, and fully connected crfs.
\newblock {\em IEEE transactions on pattern analysis and machine intelligence},
  40(4):834--848, 2017.

\bibitem{deng2009imagenet}
Jia Deng, Wei Dong, Richard Socher, Li-Jia Li, Kai Li, and Li Fei-Fei.
\newblock Imagenet: A large-scale hierarchical image database.
\newblock In {\em 2009 IEEE conference on computer vision and pattern
  recognition}, pages 248--255. Ieee, 2009.

\bibitem{deo2021multimodal}
Nachiket Deo, Eric Wolff, and Oscar Beijbom.
\newblock Multimodal trajectory prediction conditioned on lane-graph
  traversals.
\newblock In {\em 5th Annual Conference on Robot Learning}, 2021.

\bibitem{ding20201st}
Zhuangzhuang Ding, Yihan Hu, Runzhou Ge, Li Huang, Sijia Chen, Yu Wang, and Jie
  Liao.
\newblock 1st place solution for waymo open dataset challenge--3d detection and
  domain adaptation.
\newblock {\em arXiv preprint arXiv:2006.15505}, 2020.

\bibitem{Ettinger_2021_ICCV}
Scott Ettinger, Shuyang Cheng, Benjamin Caine, Chenxi Liu, Hang Zhao, Sabeek
  Pradhan, Yuning Chai, Ben Sapp, Charles~R. Qi, Yin Zhou, Zoey Yang, Aur'elien
  Chouard, Pei Sun, Jiquan Ngiam, Vijay Vasudevan, Alexander McCauley, Jonathon
  Shlens, and Dragomir Anguelov.
\newblock Large scale interactive motion forecasting for autonomous driving:
  The waymo open motion dataset.

\bibitem{gao2020vectornet}
Jiyang Gao, Chen Sun, Hang Zhao, Yi Shen, Dragomir Anguelov, Congcong Li, and
  Cordelia Schmid.
\newblock Vectornet: Encoding hd maps and agent dynamics from vectorized
  representation.
\newblock In {\em Proceedings of the IEEE/CVF Conference on Computer Vision and
  Pattern Recognition}, pages 11525--11533, 2020.

\bibitem{graham2017submanifold}
Benjamin Graham and Laurens van~der Maaten.
\newblock Submanifold sparse convolutional networks.
\newblock {\em arXiv preprint arXiv:1706.01307}, 2017.

\bibitem{hu2020probabilistic}
Anthony Hu, Fergal Cotter, Nikhil Mohan, Corina Gurau, and Alex Kendall.
\newblock Probabilistic future prediction for video scene understanding.
\newblock In {\em European Conference on Computer Vision}, pages 767--785.
  Springer, 2020.

\bibitem{hu2019learning}
Anthony Hu, Alex Kendall, and Roberto Cipolla.
\newblock Learning a spatio-temporal embedding for video instance segmentation.
\newblock {\em arXiv preprint arXiv:1912.08969}, 2019.

\bibitem{hu2021fiery}
Anthony Hu, Zak Murez, Nikhil Mohan, Sof{\'\i}a Dudas, Jeffrey Hawke, Vijay
  Badrinarayanan, Roberto Cipolla, and Alex Kendall.
\newblock Fiery: Future instance prediction in bird's-eye view from surround
  monocular cameras.
\newblock In {\em Proceedings of the IEEE/CVF International Conference on
  Computer Vision}, pages 15273--15282, 2021.

\bibitem{hu2021afdetv2}
Yihan Hu, Zhuangzhuang Ding, Runzhou Ge, Wenxin Shao, Li Huang, Kun Li, and
  Qiang Liu.
\newblock Afdetv2: Rethinking the necessity of the second stage for object
  detection from point clouds.
\newblock {\em arXiv preprint arXiv:2112.09205}, 2021.

\bibitem{izmailov2018averaging}
Pavel Izmailov, Dmitrii Podoprikhin, Timur Garipov, Dmitry Vetrov, and
  Andrew~Gordon Wilson.
\newblock Averaging weights leads to wider optima and better generalization.
\newblock {\em arXiv preprint arXiv:1803.05407}, 2018.

\bibitem{Krizhevsky2017tta}
Alex Krizhevsky, Ilya Sutskever, and Geoffrey~E. Hinton.
\newblock Imagenet classification with deep convolutional neural networks.
\newblock {\em Commun. ACM}, 60(6):84–90, may 2017.

\bibitem{lin2017focal}
Tsung-Yi Lin, Priya Goyal, Ross Girshick, Kaiming He, and Piotr Doll{\'a}r.
\newblock Focal loss for dense object detection.
\newblock In {\em Proceedings of the IEEE international conference on computer
  vision}, pages 2980--2988, 2017.

\bibitem{liu2021swinv2}
Ze Liu, Han Hu, Yutong Lin, Zhuliang Yao, Zhenda Xie, Yixuan Wei, Jia Ning, Yue
  Cao, Zheng Zhang, Li Dong, et~al.
\newblock Swin transformer v2: Scaling up capacity and resolution.
\newblock {\em arXiv preprint arXiv:2111.09883}, 2021.

\bibitem{liu2021swin}
Ze Liu, Yutong Lin, Yue Cao, Han Hu, Yixuan Wei, Zheng Zhang, Stephen Lin, and
  Baining Guo.
\newblock Swin transformer: Hierarchical vision transformer using shifted
  windows.
\newblock In {\em Proceedings of the IEEE/CVF International Conference on
  Computer Vision}, pages 10012--10022, 2021.

\bibitem{loshchilov2016sgdr}
Ilya Loshchilov and Frank Hutter.
\newblock Sgdr: Stochastic gradient descent with warm restarts.
\newblock {\em arXiv preprint arXiv:1608.03983}, 2016.

\bibitem{loshchilov2017decoupled}
Ilya Loshchilov and Frank Hutter.
\newblock Decoupled weight decay regularization.
\newblock {\em arXiv preprint arXiv:1711.05101}, 2017.

\bibitem{mahjourian2022occupancy}
Reza Mahjourian, Jinkyu Kim, Yuning Chai, Mingxing Tan, Ben Sapp, and Dragomir
  Anguelov.
\newblock Occupancy flow fields for motion forecasting in autonomous driving.
\newblock {\em IEEE Robotics and Automation Letters}, 7(2):5639--5646, 2022.

\bibitem{ngiam2021scene}
Jiquan Ngiam, Benjamin Caine, Vijay Vasudevan, Zhengdong Zhang, Hao-Tien~Lewis
  Chiang, Jeffrey Ling, Rebecca Roelofs, Alex Bewley, Chenxi Liu, Ashish
  Venugopal, et~al.
\newblock Scene transformer: A unified multi-task model for behavior prediction
  and planning.
\newblock {\em arXiv e-prints}, pages arXiv--2106, 2021.

\bibitem{oprea2020review}
Sergiu Oprea, Pablo Martinez-Gonzalez, Alberto Garcia-Garcia, John~Alejandro
  Castro-Vargas, Sergio Orts-Escolano, Jose Garcia-Rodriguez, and Antonis
  Argyros.
\newblock A review on deep learning techniques for video prediction.
\newblock {\em IEEE Transactions on Pattern Analysis and Machine Intelligence},
  2020.

\bibitem{radosavovic2020designing}
Ilija Radosavovic, Raj~Prateek Kosaraju, Ross Girshick, Kaiming He, and Piotr
  Doll{\'a}r.
\newblock Designing network design spaces.
\newblock In {\em Proceedings of the IEEE/CVF Conference on Computer Vision and
  Pattern Recognition}, pages 10428--10436, 2020.

\bibitem{shi2015convolutional}
Xingjian Shi, Zhourong Chen, Hao Wang, Dit-Yan Yeung, Wai-Kin Wong, and
  Wang-chun Woo.
\newblock Convolutional lstm network: A machine learning approach for
  precipitation nowcasting.
\newblock {\em Advances in neural information processing systems}, 28, 2015.

\bibitem{sun2020scalability}
Pei Sun, Henrik Kretzschmar, Xerxes Dotiwalla, Aurelien Chouard, Vijaysai
  Patnaik, Paul Tsui, James Guo, Yin Zhou, Yuning Chai, Benjamin Caine, et~al.
\newblock Scalability in perception for autonomous driving: Waymo open dataset.
\newblock In {\em Proceedings of the IEEE/CVF conference on computer vision and
  pattern recognition}, pages 2446--2454, 2020.

\bibitem{tan2020efficientdet}
Mingxing Tan, Ruoming Pang, and Quoc~V Le.
\newblock Efficientdet: Scalable and efficient object detection.
\newblock In {\em Proceedings of the IEEE/CVF conference on computer vision and
  pattern recognition}, pages 10781--10790, 2020.

\bibitem{wang2018eidetic}
Yunbo Wang, Lu Jiang, Ming-Hsuan Yang, Li-Jia Li, Mingsheng Long, and Li
  Fei-Fei.
\newblock Eidetic 3d lstm: A model for video prediction and beyond.
\newblock In {\em International conference on learning representations}, 2018.

\bibitem{zhao2020tnt}
Hang Zhao, Jiyang Gao, Tian Lan, Chen Sun, Benjamin Sapp, Balakrishnan
  Varadarajan, Yue Shen, Yi Shen, Yuning Chai, Cordelia Schmid, et~al.
\newblock Tnt: Target-driven trajectory prediction.
\newblock {\em arXiv preprint arXiv:2008.08294}, 2020.

\end{thebibliography}
}

\end{document}